\documentclass[]{article}
\usepackage{spconf,amsmath,graphicx}
\usepackage{cite}
\usepackage{amsfonts}
\usepackage{algorithmic}
\usepackage{graphicx}
\usepackage{textcomp}
\usepackage{xcolor}
\usepackage{graphicx}
\usepackage{amsmath}
\usepackage{amssymb}
\usepackage{booktabs}
\usepackage{svg}
\usepackage{caption}
\usepackage{subcaption}
\usepackage{multirow}
\usepackage{hyperref}
\usepackage{flushend}
\usepackage[lined,boxed,commentsnumbered]{algorithm2e}

\usepackage{times}
\usepackage{float}
\usepackage{epsfig}
\usepackage{graphicx}
\def\BibTeX{{\rm B\kern-.05em{\sc i\kern-.025em b}\kern-.08em
    T\kern-.1667em\lower.7ex\hbox{E}\kern-.125emX}}

\title{Exploring Saliency Bias in Manipulation Detection}

\name{
Joshua Krinsky\textsuperscript{1}, Alan Bettis\textsuperscript{1}, Qiuyu Tang\textsuperscript{1}, Daniel Moreira\textsuperscript{2}, Aparna Bharati\textsuperscript{1}
}
\address{\textsuperscript{1} Department of Computer Science and Engineering, Lehigh University, Bethlehem, PA, USA. \\
         \textsuperscript{2} Department of Computer Science, Loyola University, Chicago, IL, USA.}

\usepackage{fancyhdr}
\fancypagestyle{firstpage}{
    \vspace{-1.2 cm}
    \setlength{\headheight}{30mm}
    \fancyhead[l]{Copyright 2024 IEEE. Published in 2024 IEEE International Conference on Image Processing (ICIP), scheduled for 27-30 October 2024 in Abu Dhabi, United Arab Emirates. Personal use of this material is permitted. However, permission to reprint/republish this material for advertising or promotional purposes or for creating new collective works for resale or redistribution to servers or lists, or to reuse any copyrighted component of this work in other works, must be obtained from the IEEE. Contact: Manager, Copyrights and Permissions / IEEE Service Center / 445 Hoes Lane / P.O. Box 1331 / Piscataway, NJ 08855-1331, USA. Telephone: + Intl. 908-562-3966.}
    
}

\begin{document}
\maketitle

\thispagestyle{firstpage}
\fancyhf{}
\begin{abstract}
The social media-fuelled explosion of fake news and misinformation 
supported by tampered images
has led to 
growth in the development of models and datasets for image manipulation detection. 
However, existing detection methods 
mostly treat media objects in isolation, 
without considering the impact of specific manipulations on viewer perception. 
Forensic datasets are usually analyzed based on the manipulation operations and corresponding pixel-based masks, but not on the semantics of the manipulation, i.e., type of scene, objects, and viewers' attention to scene content. The semantics of the manipulation play an important role in spreading misinformation through manipulated images.
In an attempt to encourage further development of semantic-aware forensic approaches to understand visual misinformation, 
we propose a framework to analyze the trends of visual and semantic saliency in popular image manipulation datasets and their impact on detection. {\url{https://github.com/CV-Lehigh/Bias_IMD}}
\end{abstract}
\vspace{-.2cm}
\section{Introduction}
\vspace{-.2cm}
\label{sec:intro}

The increase in quality, quantity and diversity of manipulated media
has led to an increased reliance on automated visual forensics,
as human analysis has limited applicability. 
However, not all manipulated images are equally misinforming.
For efficient detection at a large scale, forensic techniques should focus more on images that have the potential for more impact on viewer perception.
In this work, we analyze image forensics datasets based on perceptual saliency, i.e., the amount of attention paid by human viewers to manipulated content within an image.
Such an analysis can be used to reduce the scale and types of images that require urgent attention and dedicated detection resources.
We hypothesize that image manipulations that catch human viewers' attention are more valuable to detect and analyze from the perspective of 
misinformation.
Gauging perceptual understanding in an image-centric manner that goes beyond just labeling images as real or fake is essential for fighting fake news and its effects. 
In this paper, we provide
the following:
\begin{itemize}
\itemsep0em 
    \item Determine the impact of saliency 
    on the human ability to detect and localize image manipulations.
    Analyze the range of visually and semantically salient splicing-based manipulations present in widely used benchmark datasets for image manipulation detection.
    
    \item Experiments showing that synthetic manipulation of saliency of image contents also lead to trends in detection performance,
    providing further evidence that more salient manipulated regions are easier to detect.

    \item Evaluate bias in the performance of manipulation detection networks based on the visual saliency of the manipulated region and the semantic change offered by the manipulation. We propose a novel method of calculating the semantic relevance of the manipulation using CLIP~\cite{clip}, a vision-language foundation model.
\end{itemize}
 \begin{figure}[!t]
     \centering
     \includegraphics[width=1.0\linewidth]{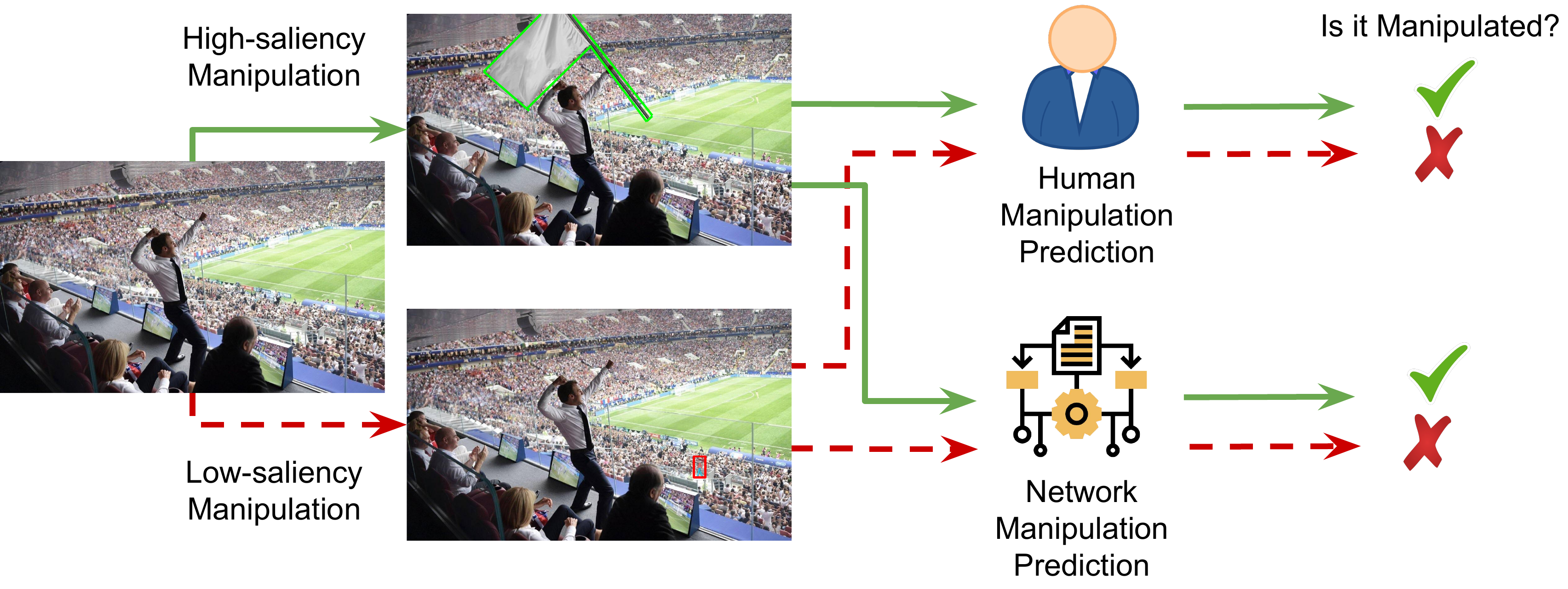}
     \vspace{-0.7cm}
     \caption{Saliency of the manipulation is an important factor in determining if a human or machine will consider an image to be manipulated.}
     \label{fig:teaser}
 \vspace{-0.5cm}
 \end{figure}

\begin{figure*}[!t]
\centering
    \begin{subfigure}{.23\linewidth}
        \includegraphics[width=1.0\linewidth]{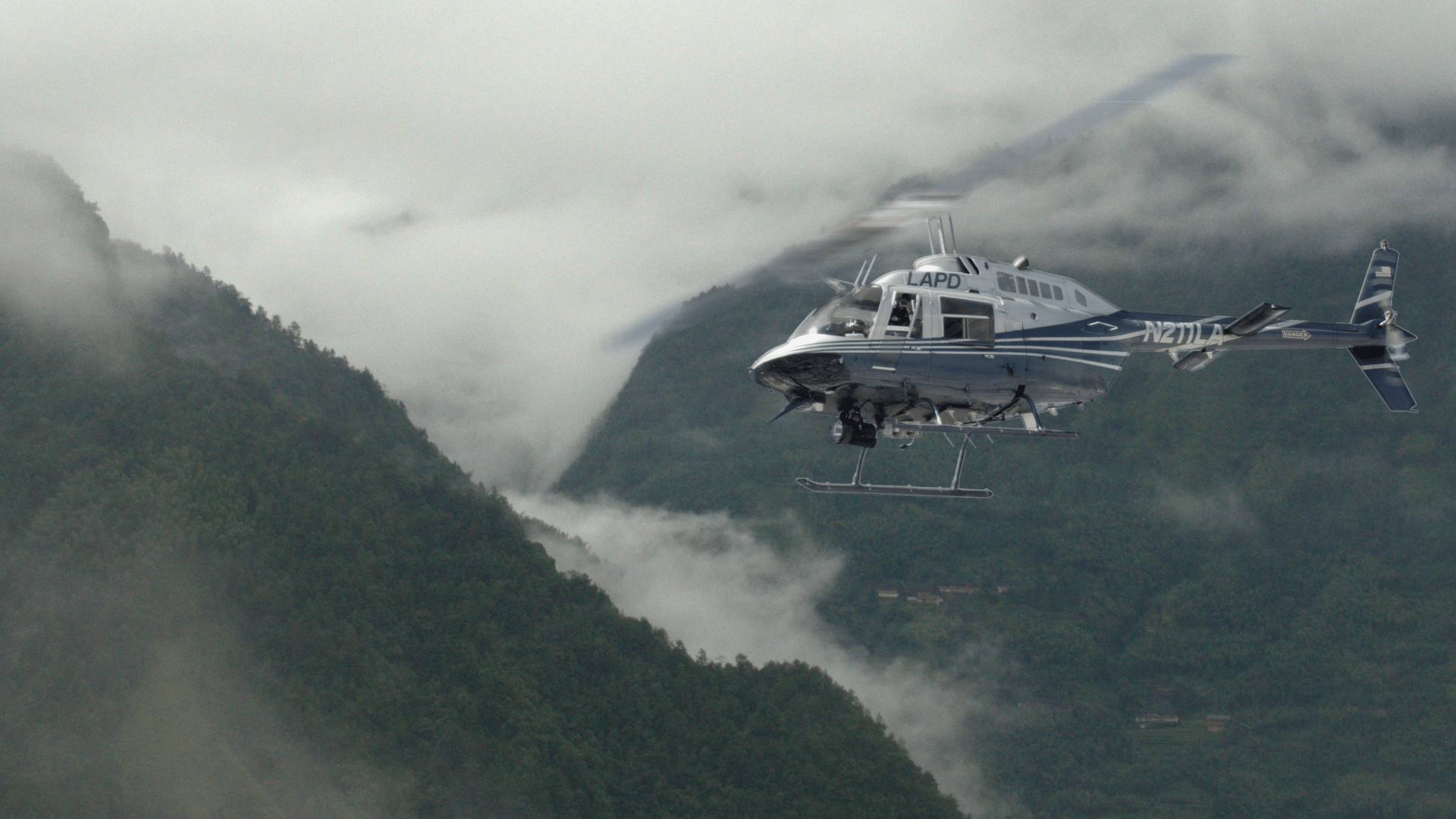}
        \caption{Image shown to the participants.}
        \label{fig:original_image}
    \end{subfigure}
    \hfill
    \begin{subfigure}{.23\linewidth}
        \includegraphics[width=1.0\linewidth]{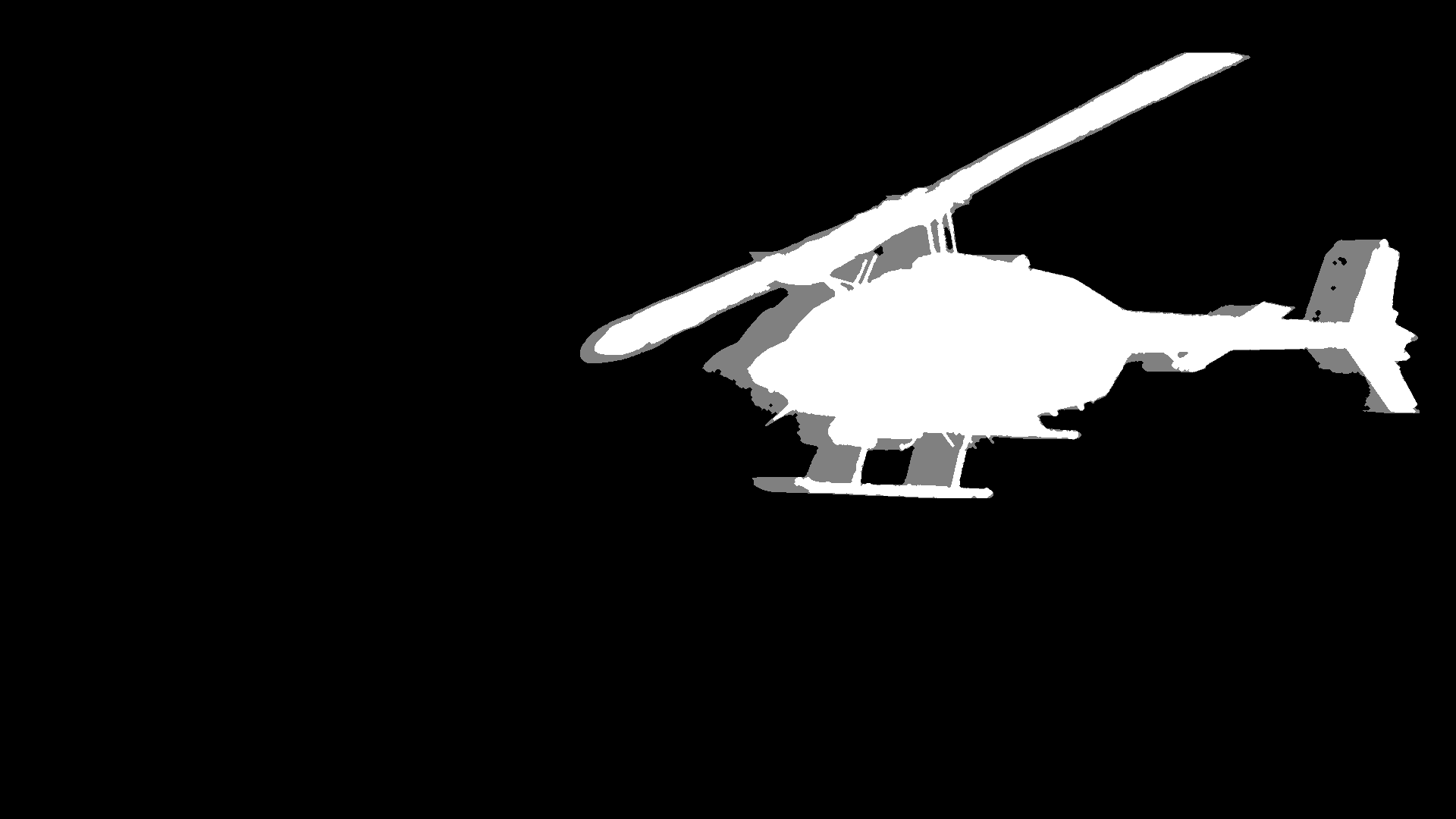}
        \caption{Ground-truth manipulation mask.}
        \label{fig:mask}
    \end{subfigure}
    \hfill
    \begin{subfigure}{.23\linewidth}
        \includegraphics[width=1.0\linewidth]{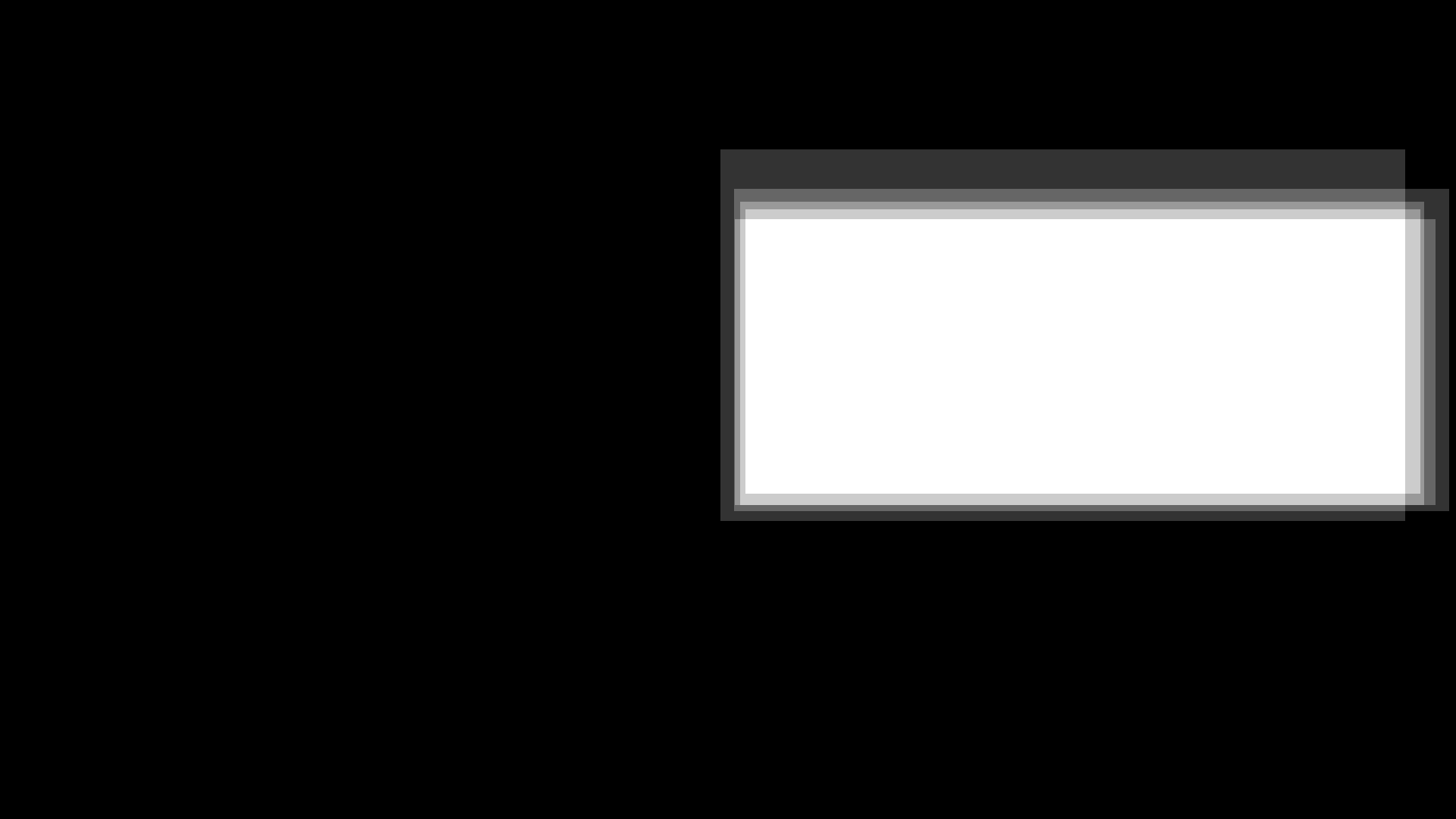}
        \caption{Region(s) perceived as most salient by the participants.}
        \label{fig:sal}
    \end{subfigure}
    \hfill
    \begin{subfigure}{.23\linewidth}
        \includegraphics[width=1.0\linewidth]{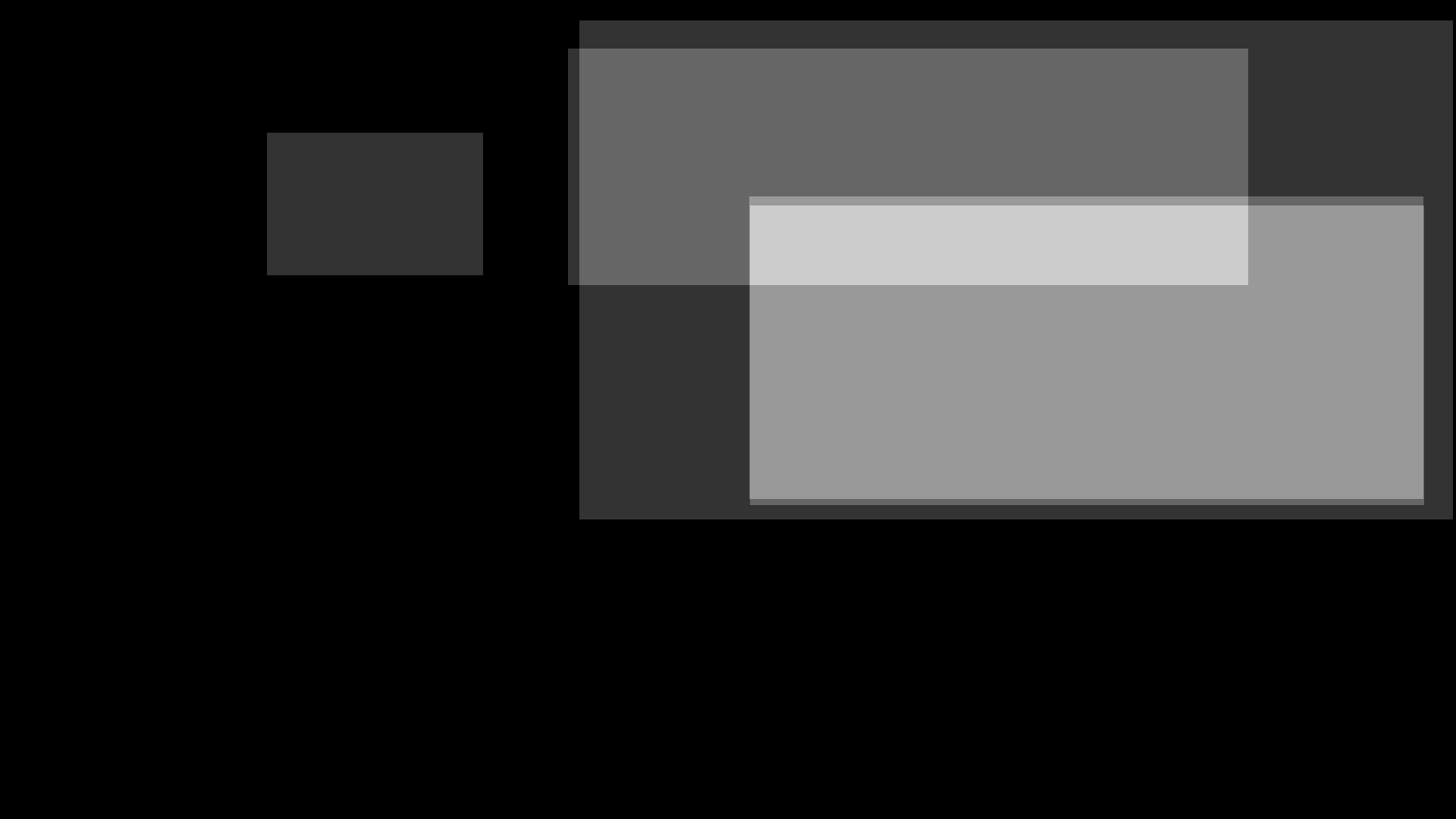}
        \caption{Region(s) perceived as manipulated by the participants.}
        \label{fig:manip}
    \end{subfigure}
    \vspace{-5pt}
   \caption{Example saliency and manipulation prediction maps obtained from the human study. Maps (b) and (c) are compared using Mean Recall to estimate the saliency of the manipulation. Maps (b) and (d) are compared using ROC to understand the accuracy of human manipulation prediction.}
   \vspace{-10pt}
\label{fig:human_study_pipeline}
\end{figure*}

\vspace{-0.48cm}
\section{Related Work}
\label{sec:related_work} 
\textbf{Human Perception and Image Manipulation Detection.}
While humans can efficiently associate semantic understanding with coherent visual scene elements~\cite{scheirer2014perceptual, rajalingham2018large}
early works in analyzing human manipulation detection ability, such as
~\cite{farid2010image}, proved that humans consistently fail to detect physical impossibilities in an image such as inconsistent shadows, perspectives, and reflections.
Recently, Nightingale et al.~\cite{nightingale2017can} and Schetinger et al.~\cite{schetinger2017humans} showed that humans are unable to distinguish real images from manipulated ones. In the respective studies, participants' could correctly classify 66\%~\cite{nightingale2017can} and 58\%~\cite{schetinger2017humans} of the presented images as real vs. fake. The former also showed that viewers couldn't locate the manipulation in the image 55\% of the time.

However, existing literature has mostly focused on evaluating manipulation detection and localization ability of humans without exploring deeper into its explanations. They do not investigate how the relevance of the manipulation or viewer attention to specific regions impacts the ability to detect it.

Saliency is a property that characterizes where people choose to focus their mental processing power. Visual saliency plays a key role in biasing human attention~\cite{itti1998model} and processing visual stimuli~\cite{borji2013stands}.
Perception scientists have studied how salient aspects of an image impact human performance on a range of perceptual tasks, such as user engagement~\cite{mccay2012saliency}, category learning~\cite{hammer2015impact}, and distinguishing between real and fake image~\cite{nightingale2017can, korshunov2020deepfake}. 

Rensink et al.'s seminal work, ``To See or Not To See''~\cite{rensink1997see}, studied the human ability to detect changes in scenes that belonged to one of two saliency-related categories, central interest and marginal interest. In the study, participants would swap between two images, one original and one manipulated, until they found the difference between the two. 
The authors concluded that visual perception of change of an object occurs only when that object is given focused attention.
In this paper, we perform a human study that investigates
how the saliency of manipulations in real-world images affects the users' ability to detect manipulations. 
Unlike the study in~\cite{rensink1997see},
to emulate a more realistic online scenario, we show manipulated images without an original reference images 
and ask the participants 
to 
annotate the regions they pay 
attention to and those they think were manipulated. 

\noindent\textbf{Learning-based Manipulation Detection Models.}
With the advent of deep learning, 
data-driven solutions for manipulation detection were developed.
In recent years, Convolutional Neural Networks (CNNs)
have been applied to the manipulation detection task~\cite{Wu2019ManTraNet, PSCC-Net, wu2018busternet, OSN}. These networks learn patterns that distinguish pristine and manipulated images from human-labeled data.
Most notably, the Local Anomaly Detection Network (LADN) architecture proposed by ManTra-Net~\cite{Wu2019ManTraNet}
was designed to mimic the human decision-making process. 
Similarly, the authors of PSCC-Net~\cite{PSCC-Net} got inspired by how people solve tasks going from a coarse to fine image analysis to detect manipulations.
Thus, certain biases originating from data curation can influence learning-based models trained on this data. 
Other models
have explored attention networks for context-aware manipulation detection~\cite{ren2023multi} and fine-grained hierarchical manipulation classification~\cite{guo2023hierarchical}.
This work investigates biases observed in image manipulation detection models and compares them with those observed in humans.

\vspace{-10pt}
\section{Human Saliency}
\label{sec:human_study}
\vspace{-3pt} 
To understand
how the saliency of the manipulations is related to peoples' ability to accurately
spot them, i.e., if people are better at detecting manipulations when they are within the salient region of the image, 
we set up a user study.
Identifying saliency bias within human manipulation detection can help develop tools to help humans spot less salient manipulations.

In the study, participants looked at 130 spliced images from the Korus' Realistic Tampering (RT) dataset~\cite{korus2016multi,korus2016evaluation} containing well-crafted manipulations proven to be challenging for
manipulation detection and localization 
networks~\cite{mayer2020exposing}.
For each image, the participants completed 2 tasks. The first task was
to place bounding boxes
on objects or regions that looked the most attention-grabbing.
This question identified the salient regions of the image. 
For the second task, the participants placed bounding boxes 
on objects or regions they believed to be manipulated, if any. 
The order of these questions is important. We wanted to first understand what a human focuses on (saliency), and then ask specifically if they find something manipulated in the image. Since saliency is a more general visual concept, the saliency question being presented first does not shift the participants' opinion of manipulation because saliency can be assessed from any image regardless of whether it is manipulated. Additionally, we chose to ask these questions to the same participants instead of conducting two studies for collecting the two types of responses. There can be variations in the human detection ability and visual reasoning. Recording both in the same session maintains consistency between saliency and manipulation data. 

The study was created by the authors~\footnote{{\url{http://pnz.aca.mybluehost.me}}} and participants were recruited using Prolific~\footnote{{\url{https://www.prolific.com}}}, a crowdsourcing platform which ensures highly qualified and vetted participants~\cite{douglas2023data}.
Each image was reviewed 5 times and 650 responses were recorded in total from 65 individuals. The participants were compensated \$2.00 per survey consisting of 10 images each. 
The bounding boxes recorded from all the participants' responses were combined to create a (i) human saliency map~\ref{fig:sal} 
and the corresponding (ii) manipulation prediction mask~\ref{fig:manip}. The final masks for each image contain a higher weight or confidence for pixel locations where boxes provided by multiple participants overlapped.
Both the human saliency and manipulation prediction masks are then compared to its respective ground-truth pixel-wise manipulation mask~\ref{fig:mask}.
This comparison 
reveals 
how salient a manipulation is to the participants, and how well they could localize the manipulation (used to also determine detection performance, as correct localization implies correct detection implicitly).
To obtain the saliency score of each image manipulation, the saliency mask was compared with the ground truth using pixel-wise Mean Recall, which considers both accuracy and group agreement.
\begin{figure}[!t]
\centering
\includegraphics[width=0.9\linewidth]{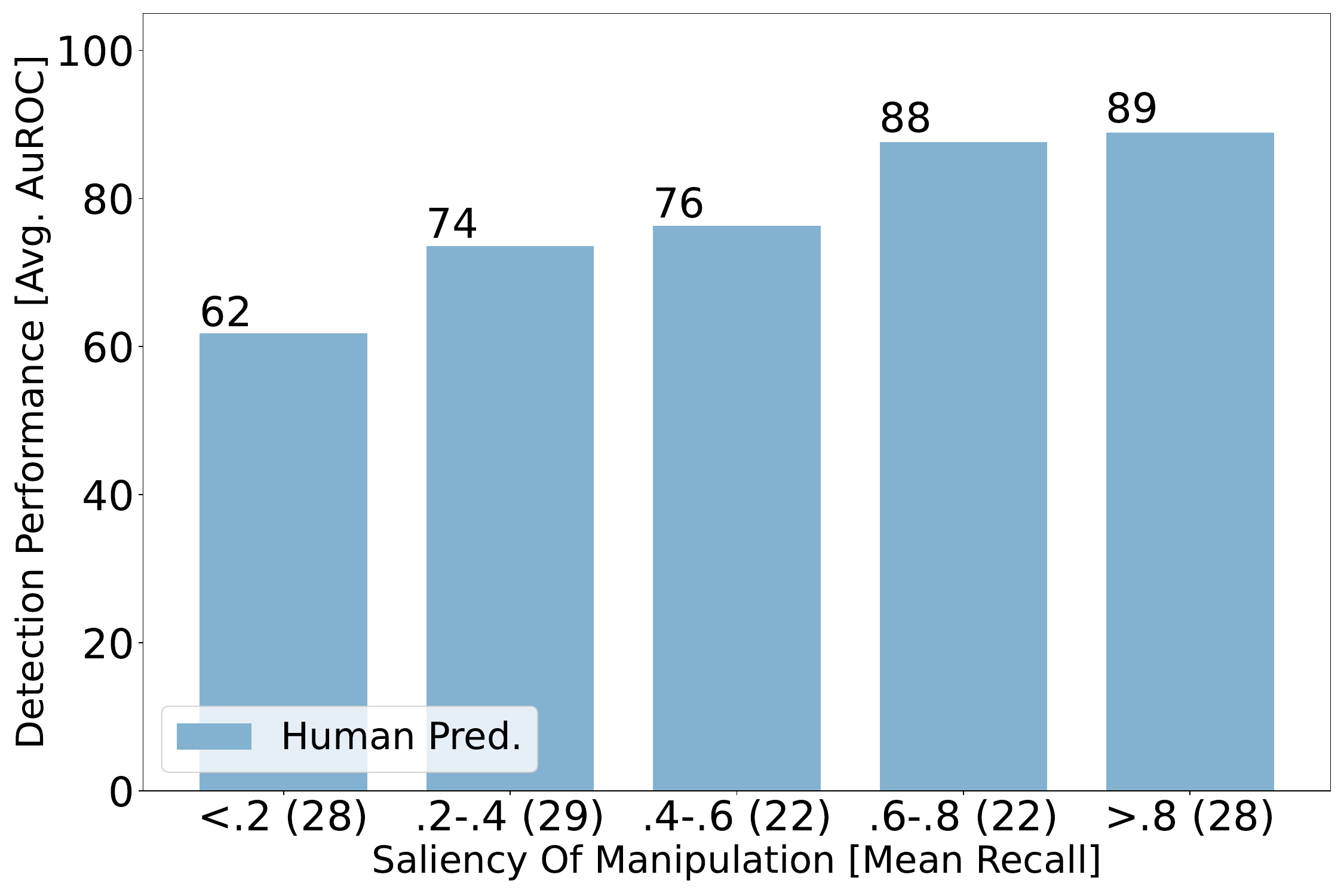}
\vspace{-10pt}
\caption{Detection performance (AuROC) from human participants for five saliency 
levels over 
the RT dataset (number of images in each group in parentheses).
}
\label{fig:human_results}
\vspace{-.7cm}
\end{figure}

The images used in the study were then divided into five groups depending on the saliency score of their manipulations. We obtained a fairly even distribution of images in the five groups, implying that
the RT dataset contains splicing manipulations evenly spread across different levels of saliency.
To understand if the human performance worsens when the less salient image regions were manipulated, we computed the detection performance of human participants for the images in each group. For each image within a group, the manipulation prediction mask was compared to the ground truth manipulation mask and the Area under the Receiver Operating Characteristics curve (AuROC) for pixel-wise manipulation classification was used.
The average AuROC for each level of saliency is reported in Fig.~\ref{fig:human_results} and shows a correlation between the saliency of the manipulations and how well people can localize them.
The detection performance
increases consistently through groups, from around $0.62$ in the first group to $0.89$ in the final group,
showcasing that the saliency of the manipulations can bias people's ability to properly localize manipulations.

\vspace{-10pt}
\section{Machine Experiments}
\label{sec:experiment}
\vspace{-0pt}

Results from the human study lead us to formulate two questions:
first, do manipulation detection networks reflect similar biases as seen in humans, and are the trends similar in other image manipulation datasets?
And second, 
will increasing the saliency of these manipulations correlate to improved human detection performance? Answering the latter also further verifies if saliency was a contributor to the performance trends over other variables such as content in the image, quality, etc. 
This section describes the experiments conducted 
to answer the two  
research questions. 

\begin{figure}[!t]
  \centering
    \includegraphics[width=0.9\linewidth]{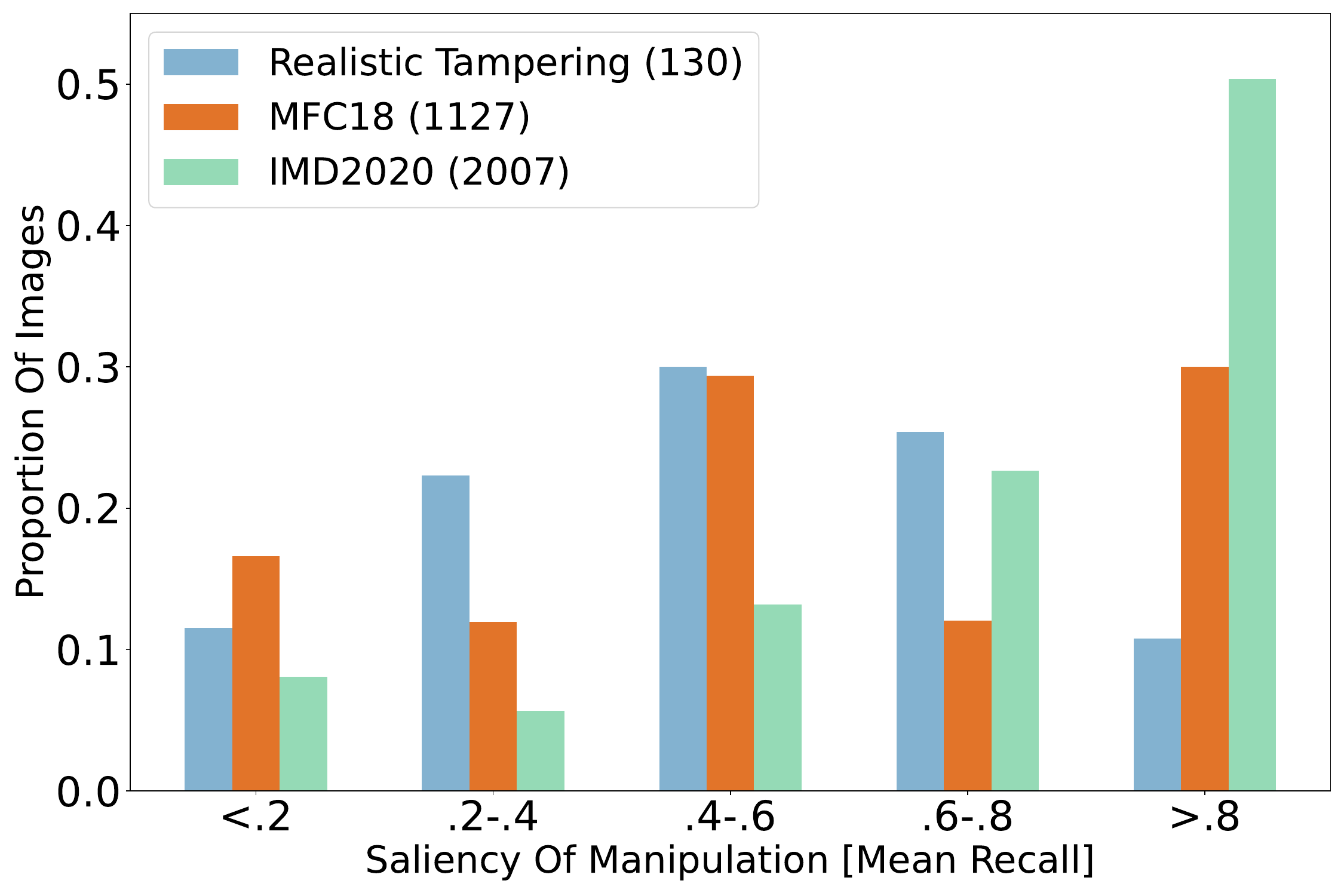}
  \vspace{-10pt}
  \caption{Proportion of the dataset in each saliency group.
  }
  \label{fig:saliency_distribution}
\vspace{-0.5cm}
\end{figure}

\noindent \textbf{Datasets Used.} For enabling comparison between human performance and that of detection algorithms, the first adopted
dataset
refers to 
the same subset of 130 images 
from the Korus' 
RT dataset as used in the human study (Sec.~\ref{sec:human_study}).

We use two other forgery datasets, which enable larger-scale automated analyses. 
The 2018 Media Forensics Challenge dataset (MFC18)~\cite{guan2019mfc} was released by the American National Institute of Standards and Technology (NIST), and 
contains 
several forgery instances (more examples than the RT dataset). 
It contains multiple manipulations on the same image,
which is a typical technique utilized by forgers to improve realism and hide detectability.
Due to input resolution constraints, we select 1127 images from MFC18 with sum of dimensions up to 4K pixels and at least one manipulation that can change the semantic meaning of the image (e.g., copy-move, splicing, and inpainting).
The second dataset, IMD2020~\cite{novozamsky2020imd2020}, contains 2007 images sourced from the \url{r/photoshopbattles} subreddit. 
This crowdsourced database provides the benefit of massive variance in image acquisition conditions and content.
However, as the main goal of the community is to create obvious, funny, and satirical manipulations, the forgeries may not be well disguised.

\begin{figure*}
\vspace{-20pt}
    \centering
    \includegraphics[width=1.0\linewidth]{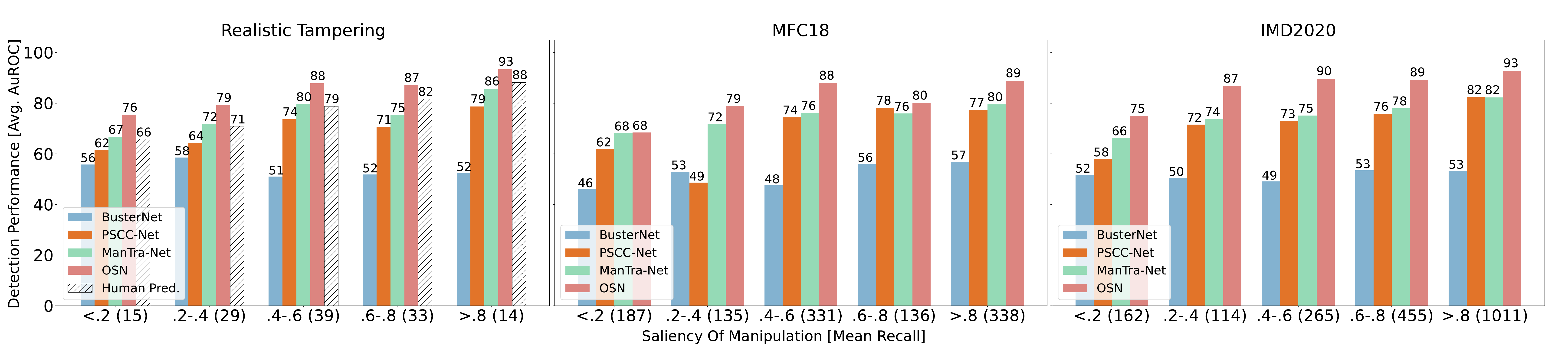}
    \vspace{-15pt}
    \caption{Manipulation detection and localization (average AuROC) results on the 3 datasets. The first plot additionally shows efficacy of manipulated region predictions collected from the human study.}
    \label{fig:performance_distribution}
    \vspace{-10pt}
\end{figure*}

\begin{figure*}[!th]
\begin{center}
\includegraphics[width=0.8\linewidth]{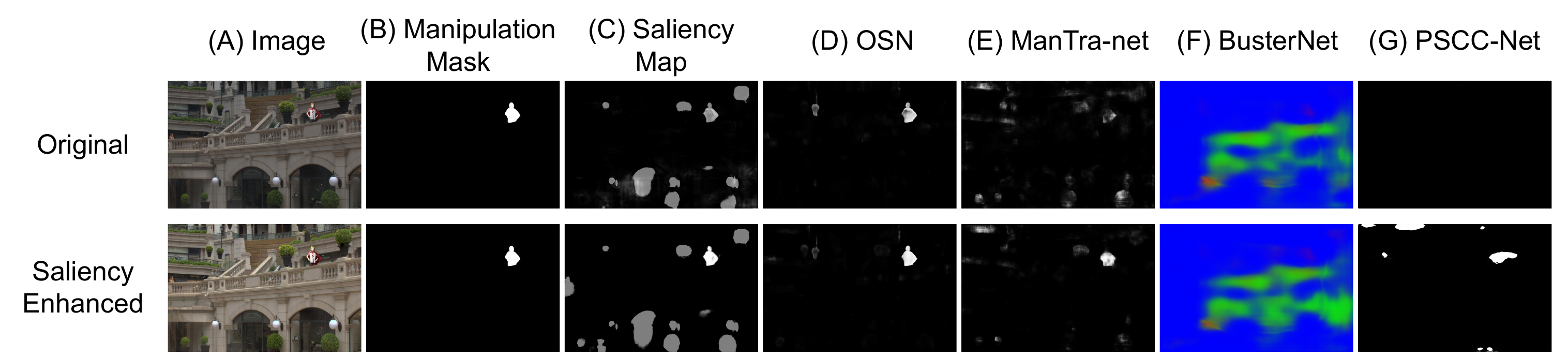}
\end{center}
\vspace{-10pt}
   \caption{
   Top: tampered image and its predictions from different networks. Bottom: saliency enhanced tampered image and its predictions.
   (D) - (G) prediction from each of the manipulation detection and localization networks~\cite{OSN, Wu2019ManTraNet, wu2018busternet, PSCC-Net}.
   }
\label{fig:saliency_enhance_example}
\vspace{-10pt}
\end{figure*}

\noindent\textbf{Experiment~1.}
\label{sec:experiment1}
~\textit{Dataset Visual Saliency Bias Estimation.}
The process of collecting human annotations
cannot be extrapolated for every dataset,  
as it is time-consuming and expensive.
In order to get robust saliency predictions for the larger datasets, we utilize two object-based saliency prediction networks, U$^2$-Net~\cite{qin2020u2} and R$^3$Net~\cite{deng2018r3net}. Both networks generate a saliency prediction within the range $[0, 1]$ for each pixel in a given image.
Averaging the two saliency maps helps build robustness and accommodate scenarios where a single network fails.
Since saliency is subjective, it is important to consider multiple opinions when deciding the saliency, as with human responses. The combined saliency map from U$^2$-Net and R$^3$Net is compared to the ground truth manipulated mask using Mean Recall. Based on the mean recall score, images are placed into one of five saliency groups.

The distribution of saliency of manipulations for the RT~\cite{korus2016evaluation}, MFC18~\cite{guan2019mfc} and IMD2020~\cite{novozamsky2020imd2020} datasets is shown in Fig.~\ref{fig:saliency_distribution}. Compared to the saliency distribution generated by the human study, the RT dataset is less uniformly distributed. However, it is 
still more uniformly distributed than MFC18 and IMD2020. After splitting images into the proper saliency groups, we calculate the average AuROC (detection performance) for each group across popular manipulation detection and localization models: PSCC-Net~\cite{PSCC-Net}, OSN~\cite{OSN}, BusterNet~\cite{wu2018busternet}, and ManTra-net~\cite{Wu2019ManTraNet}. AuROC is commonly used to evaluate manipulation detection models and allows for comparison of model performance across datasets.
Evaluating the detection performance for images in each saliency group individually can help determine how much visual saliency has an effect on manipulation detection.

\noindent \textbf{Experiment~2.}
~\textit{Impact of Varying Saliency on Manipulation Detection.}
As observed so far, saliency of the manipulated region plays a role in a model's ability to detect the manipulation. 
If saliency bias exists within manipulation detection networks, the average detection performance for each saliency group should increase as saliency of manipulations in images of the group increases. Essentially, when a manipulation becomes more salient, models should detect the manipulation more accurately.
Saliency of difficult-to-detect manipulated regions in a given image can be increased using a saliency-guided image manipulation network~\cite{mejjati2020look, chen2019guide, miangoleh2023realistic}. 
Such networks attempt to modify color, contrast, and saturation, while avoiding changes that can alter semantic interpretation of the image.
Evaluating average detection performance of saliency enhanced manipulations across multiple networks can provide additional evidence that saliency is a biasing factor for manipulation detection.

\begin{figure*}[h]
\vspace{-20pt}
\centering
    \includegraphics[width = 0.9\linewidth]{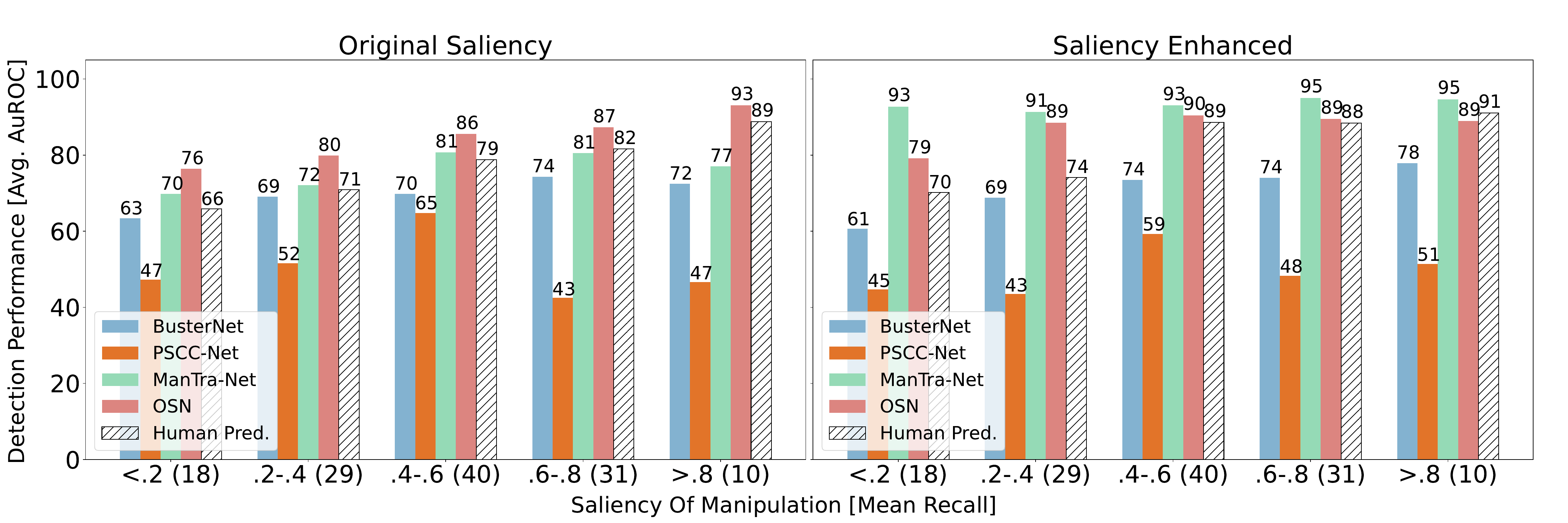}
    \vspace{-5pt}
    \caption{SaGIM~\cite{chen2019guide} allows for the manipulated region to become more salient and results in increased detection performance. Results shown on RT dataset (2 outlier images removed). 
    Left: Average AuROC of resized (to fit SaGIM requirements) RT images for each saliency of manipulation group. Right: Average AuROC of saliency enhanced RT images.}
    \label{fig:korus_sal_enhance_results}
\label{fig:korus_saliency_enhance}
\vspace{-0.2cm}
\end{figure*}

To understand if the detectability of manipulated images improves upon artificially guiding attention towards relevant areas, we employ the GAN-based method proposed in~\cite{chen2019guide}.
Upon providing two images, an RGB image and a binary attention mask, the Saliency-Guidance Image Manipulation (SaGIM) network~\cite{chen2019guide} attempts to modify the RGB image such that the regions highlighted in the attention mask are more salient. 
Average detection performance for each manipulation detection network is calculated for the saliency enhanced variants of the images in each manipulation saliency group. 
The performance before and after saliency-guided image manipulation are presented in Fig.~\ref{fig:korus_saliency_enhance}.

Additionally, another human study was conducted with images post saliency enhancement, following the same protocol as described in Sec.~\ref{sec:human_study},
to evaluate if saliency adjustment helps people detect manipulations. 
130 images received responses from 5 participants each and their overall detection performance is shown in Fig.~\ref{fig:korus_saliency_enhance}.

\begin{table*}
\footnotesize
\caption{Average detection performance per saliency group for various manipulation detection networks. 
The result shows a decrease in performance variation between low and high salient manipulations after saliency enhancement.}
\vspace{-0.2cm}
    \centering
    \begin{tabular}{l|c|c|cccc|cccc}
    \multicolumn{1}{c}{} & \multicolumn{1}{c}{} & \multicolumn{1}{c}{} & \multicolumn{4}{c}{\textbf{Original Saliency}} & \multicolumn{4}{c}{\textbf{Saliency Enhanced}} \\
    \cmidrule(rl){4-7} \cmidrule(rl){8-11}
    {Dataset} & {Partition} & {count} & {BusterNet} & {PSCC-Net} & {ManTra-Net} & {OSN} & {BusterNet} & {PSCC-Net} & {ManTra-Net} & {OSN}\\
    \midrule
    \multirow{5}{*}{IMD2020} & $<$ .2 & 179 &     0.53 & 0.57 & 0.65 & 0.71 & 0.55 & 0.72 & 0.90 & 0.79\\
                            & .2 $-$ .4 & 118 &  0.59 & 0.59 & 0.71 & 0.82 & 0.61 & 0.74 & 0.90 & 0.85\\ 
                           & .4 $-$ .6 & 270 &  0.65 & 0.59 & 0.72 & 0.86 & 0.67 & 0.73 & 0.91 & 0.89\\
                          & .6 $-$ .8 & 431 &  0.68 & 0.56 & 0.75 & 0.86 & 0.70 & 0.71 & 0.90 & 0.89\\
                         & $>$ .8 & 1008 &    0.72 & 0.59 & 0.79 & 0.90 & 0.73 & 0.72 & 0.93 & 0.91\\
    \bottomrule
    \end{tabular}
    \label{tab:SE_performance}
    \vspace{-15pt}
    \end{table*}
    
\vspace{-5pt}
\section{Results}
\vspace{-0pt}
\label{sec:results}

\textbf{Is there performance bias for manipulation detection algorithms based on the visual saliency of the manipulated region?}
Based on Experiment 1, discussed in Section~\ref{sec:experiment1},
a similar trend of results across multiple networks provide tangible evidence of saliency bias. 
The results from the evaluation show a clear increase in performance as the manipulations get more salient across all evaluated datasets and using all models except BusterNet (see Fig. \ref{fig:performance_distribution}). The failure can be attributed to its inability to handle larger images
This clear performance gap between saliency groups similar to the one seen in human performance (Fig.~\ref{fig:human_results}) indicates that saliency is a clear factor in detectability of manipulations. The larger number of low-salient manipulations combined with detection performance bias may explain why RT is considered such a difficult dataset for many manipulation detection networks~\cite{mayer2020exposing, agrawal2022sisl}.

\noindent\textbf{Does varying the saliency of the manipulated region change the detection performance for machines and humans?}
Experiment 2 was aimed to test if there is an increase in average detection performance as we increased the saliency of the manipulated regions in images, resulting in a shrinking of the performance gap between low salient images and high salient images. This direct relation between manipulation saliency and detection performance (Table \ref{tab:SE_performance} and Fig. \ref{fig:korus_sal_enhance_results}) reinforces evidence for our hypothesis that the more salient a manipulation, the more accurate its detection is by both humans 
and models.

A possible source of improvement in algorithmic detection performance is networks detecting the manipulation performed by the SaGIM network. However, if the networks were detecting the global changes made by SaGIM, the detection performance scores would be low by virtue of increased false positive rate, i.e., 
where the models predicted a pixel as manipulated when it was not indicated as manipulated in the original ground truth mask. Additionally, BusterNet which previously failed to score well (Fig.~\ref{fig:performance_distribution}) improves in detection performance due to resizing (downsampling as per input dimension requirements of SaGIM) improving its ability. However, saliency enhancement led to minor improvements in its performance. Conversely, the resizing caused PSCC-Net to perform worse, also reported in the original paper~\cite{PSCC-Net} and while saliency enhancement has a positive impact, it is not enough to surpass its performance on the unprocessed images.
Fig.\ref{fig:saliency_enhance_example} illustrates and summarizes the impact of using saliency re-attention over a manipulated image. Finally, a marginal decline in detection performance is observed for highly salient images following saliency enhancement. This phenomenon arises from the inherent challenge of augmenting the saliency of regions already characterized by high salience. Enhancement attempts may inadvertently introduce manipulations that diminish the saliency of the region, consequently reducing the overall performance. 
\vspace{-5pt}
\section{Semantic Relevance of Visual Saliency}
\vspace{-0pt}

The semantics within an image that a human focuses on can also represent saliency and can directly impact the perceived message. If an image region is not visually salient, it may not contribute to the interpretation of the image by a human viewer and can lead to misinformation. 
The semantic description of a scene can be obtained by either asking human participants (expensive and difficult to standardize) or using a multimodal foundation model, that embeds both images and semantic descriptions, originally provided by humans, into the same latent space. 

\begin{figure}[t]
    \centering
    \includegraphics[width=\linewidth]{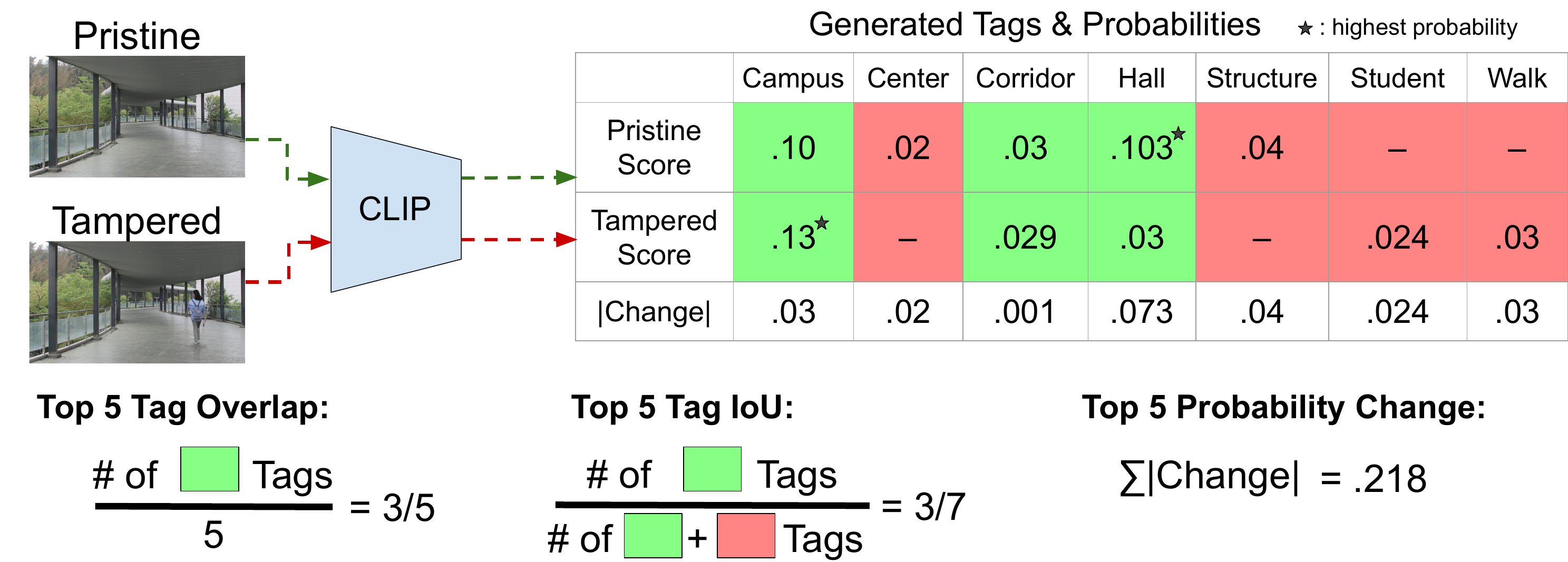}
    \caption{
    The CLIP model generates the probability of relevance for every word in the corpus based on an image. The tags shared by the pristine and tampered images are in green, and the rest in red. We calculate the percentage of common semantic concepts when considering top 5 relevant words from the pristine and manipulated image predictions.}
    \label{fig:CLIP}
    \vspace{-15pt}
\end{figure}

To figure out the effect of saliency of a manipulation on the semantic interpretation of images, we use a pre-trained Contrastive Language-Image Pretraining (CLIP)~\cite{clip} model
to compare the important semantic elements within a scene
before and after spliced edits or manipulation. 
CLIP has been used for evaluation of various high-level tasks such as image captioning~\cite{hessel2021clipscore} and image reconstruction~\cite{takagi2022psm}. In a similar vein, we employ it to evaluate the semantic change caused by manipulation. 
Given an image and a text corpus, in our case, a dictionary of nouns~\footnote{{\url{https://en.wiktionary.org}}}, the model tries to relate the semantic content in images with text and returns a list of words relevant to the scene and their probabilities.
By applying CLIP to both pristine image and their manipulated variant, we can investigate the correlation of the visual manipulation and its saliency to what machine algorithms find relevant in a scene. The exact metrics compare the words predicted with the highest relevance, i.e., probability, for both pristine and manipulated images and are explained with an example in Fig~\ref{fig:CLIP}.

\begin{figure}[t]
\begin{center}
\includegraphics[width=0.9\linewidth]{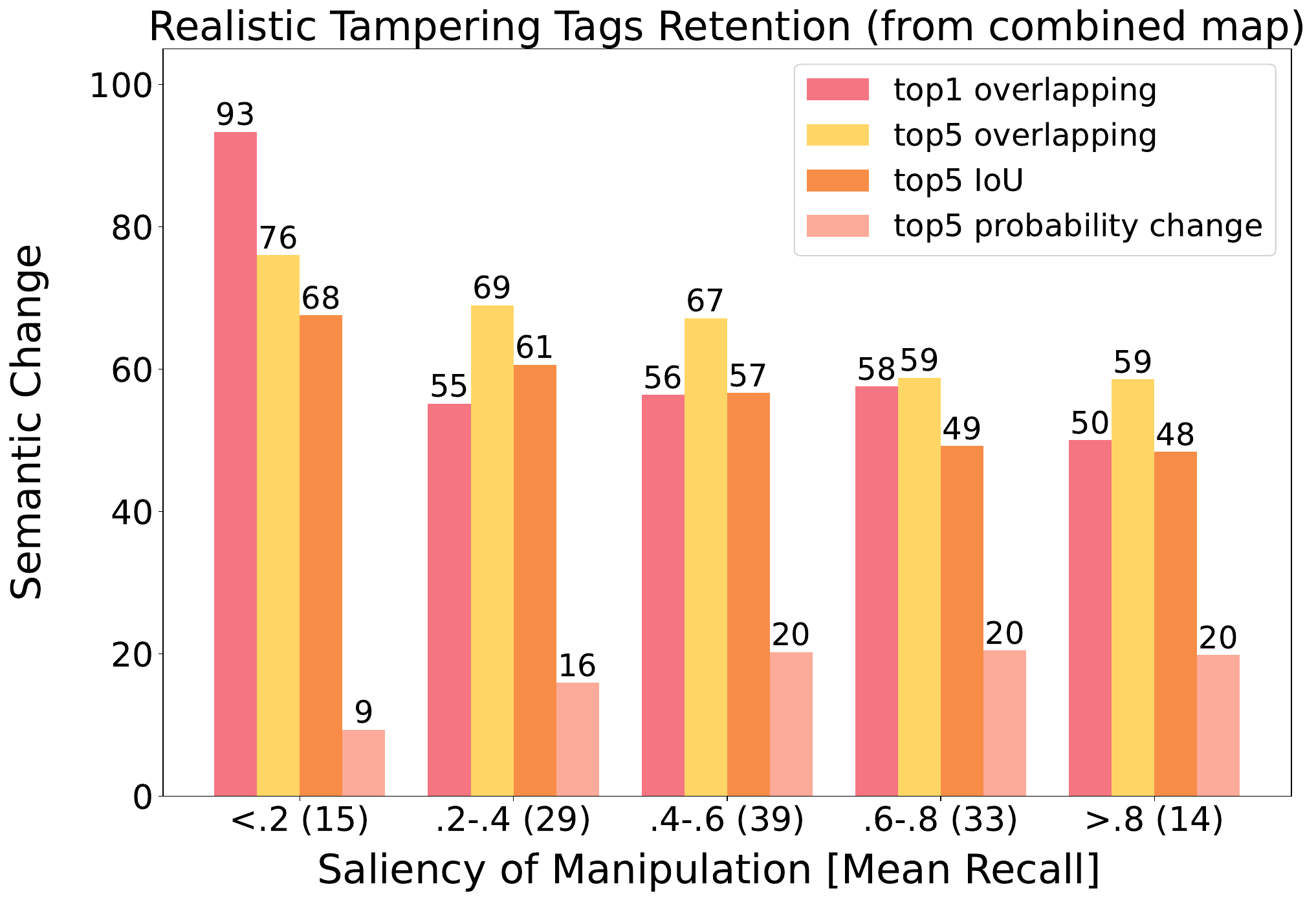}
\end{center}
    \vspace{-15pt}
       \caption{Semantic change metrics for image groups created using automated saliency estimation splits for RT dataset.
       High overlap scores indicate more similarity between original and tampered images, hence the manipulation lends low semantic change, and vice versa. 
       }
    \label{fig:PSM_semantic_saliency}
    \vspace{-15pt}
\end{figure}

Semantic change is analyzed using
the aggregated change in the predicted tag lists and probabilities yielded by the pristine and tampered versions of an image, for 5 trials (since the model is stochastic in its predictions). 
Specifically, we use top1 overlap, top5 overlap, top5 IoU, and top5 probability change as metrics (reported in Fig.~\ref{fig:PSM_semantic_saliency} for RT and IMD2020 datasets for manipulations with varying saliency). 
If the pristine and tampered tag list has a top1 overlap score of 0, the primary semantic meaning was changed by the manipulation. Similarly, if the top5 IoU and top5 overlap are low, it indicates the manipulated region greatly changes the semantic meaning. 
Top5 probability change calculates the sum of change in probability of the top 5 tags for each image.

\noindent\textbf{Do visual saliency biases also relate to semantic difference interpreted by general purpose vision models?}
Fig.~\ref{fig:PSM_semantic_saliency} shows that on average, the higher the saliency of manipulation, the lower the overlap metric scores. 
The top1 overlap is initially $0.93$, but decreases to $0.50$ by the final saliency group. The decrease in overlap metrics implies that there is higher perceivable semantic change as the manipulated region gets more salient.
Similarly, the top 5 probability change metric increases, starting from 9\% probability change for the lowest salient set to 20\% for the highest. The increase in probability change metric with increase in saliency of the manipulation shows that the higher the saliency of the manipulation, the greater the semantic change from the manipulation.

\vspace{-10pt}
\section{Conclusion}
\label{sec:conclusion}
\vspace{-10pt}
This paper formally identifies saliency of the manipulation as a factor in its detectability. 
Manipulations in two of the three investigated datasets are diverse with regard to saliency and IMD2020 has generally high salient manipulations. Our results conclude that the saliency of the manipulation is an important factor in changing the semantic meaning of the image and the ability for both people and networks to localize it. Additionally, we show that increasing the saliency of the manipulated region with 
tools such as~\cite{chen2019guide, mejjati2020look, miangoleh2023realistic} results in an increased  detection performance
for both humans and detection networks, reinforcing the hypothesis that saliency affects detection performance of manipulated images.  
\bibliographystyle{IEEEbib}
\bibliography{refs_short}

\begin{thebibliography}{10}

\bibitem{clip}
Alec Radford et~al.,
\newblock ``Learning transferable visual models from natural language supervision,''
\newblock in {\em ICML}, 2021, pp. 8748--8763.

\bibitem{scheirer2014perceptual}
Walter Scheirer et~al.,
\newblock ``Perceptual annotation: Measuring human vision to improve computer vision,''
\newblock {\em IEEE TPAMI}, vol. 36, pp. 1679--1686, 2014.

\bibitem{rajalingham2018large}
Rishi Rajalingham et~al.,
\newblock ``Large-scale, high-resolution comparison of the core visual object recognition behavior of humans, monkeys, and state-of-the-art deep artificial neural networks,''
\newblock {\em Journal of Neuroscience}, vol. 38, no. 33, pp. 7255--7269, 2018.

\bibitem{farid2010image}
Hany Farid and Mary Bravo,
\newblock ``Image forensic analyses that elude the human visual system,''
\newblock in {\em SPIE Media forensics and security II}, 2010, vol. 7541, pp. 52--61.

\bibitem{nightingale2017can}
Sophie Nightingale, Kimberley Wade, and Derrick Watson,
\newblock ``Can people identify original and manipulated photos of real-world scenes?,''
\newblock {\em Springer Cognitive research: principles and implications}, vol. 2, no. 1, pp. 1--21, 2017.

\bibitem{schetinger2017humans}
Victor Schetinger et~al.,
\newblock ``Humans are easily fooled by digital images,''
\newblock {\em Elsevier Computers \& Graphics}, vol. 68, pp. 142--151, 2017.

\bibitem{itti1998model}
Laurent Itti, Christof Koch, and Ernst Niebur,
\newblock ``A model of saliency-based visual attention for rapid scene analysis,''
\newblock {\em IEEE TPAMI}, vol. 20, pp. 1254--1259, 1998.

\bibitem{borji2013stands}
Ali Borji, Dicky Sihite, and Laurent Itti,
\newblock ``What stands out in a scene? a study of human explicit saliency judgment,''
\newblock {\em Elsevier Vision Research}, vol. 91, pp. 62--77, 2013.

\bibitem{mccay2012saliency}
Lori McCay-Peet, Mounia Lalmas, and Vidhya Navalpakkam,
\newblock ``On saliency, affect and focused attention,''
\newblock in {\em ACM CHI}, 2012, pp. 541--550.

\bibitem{hammer2015impact}
Rubi Hammer,
\newblock ``Impact of feature saliency on visual category learning,''
\newblock {\em Frontiers in Psychology}, vol. 6, pp. 451, 2015.

\bibitem{korshunov2020deepfake}
Matthew Groh et~al.,
\newblock ``Deepfake detection by human crowds, machines, and machine-informed crowds,''
\newblock {\em PNAS}, vol. 119, no. 1, 2022.

\bibitem{rensink1997see}
Ronald Rensink, Kevin O'Regan, and James Clark,
\newblock ``To see or not to see: The need for attention to perceive changes in scenes,''
\newblock {\em Psychological Science}, vol. 8, no. 5, pp. 368--373, 1997.

\bibitem{Wu2019ManTraNet}
Yue Wu, Wael Abd-Almageed, and Prem Natarajan,
\newblock ``Mantra-net: Manipulation tracing network for detection and localization of image forgeries with anomalous features,''
\newblock in {\em IEEE/CVF CVPR}, 2019.

\bibitem{PSCC-Net}
Xiaohong Liu et~al.,
\newblock ``Pscc-net: Progressive spatio-channel correlation network for image manipulation detection and localization,''
\newblock {\em IEEE TCSVT}, vol. 32, pp. 7505--7517, 2022.

\bibitem{wu2018busternet}
Yue Wu, Wael Abd-Almageed, and Prem Natarajan,
\newblock ``Busternet: Detecting copy-move image forgery with source/target localization,''
\newblock in {\em ECCV}, 2018, pp. 168--184.

\bibitem{OSN}
Haiwei Wu et~al.,
\newblock ``Robust image forgery detection against transmission over online social networks,''
\newblock {\em IEEE TIFS}, vol. 17, no. 1, pp. 443--456, 2022.

\bibitem{ren2023multi}
Ruyong Ren et~al.,
\newblock ``Multi-scale attention context-aware network for detection and localization of image splicing: Efficient and robust identification network,''
\newblock {\em Springer Applied Intelligence}, pp. 1--20, 2023.

\bibitem{guo2023hierarchical}
Xiao Guo et~al.,
\newblock ``Hierarchical fine-grained image forgery detection and localization,''
\newblock in {\em IEEE/CVF CVPR}, 2023, pp. 3155--3165.

\bibitem{korus2016multi}
Pawe{\l} Korus and Jiwu Huang,
\newblock ``Multi-scale analysis strategies in prnu-based tampering localization,''
\newblock {\em IEEE TIFS}, vol. 12, no. 4, pp. 809--824, 2016.

\bibitem{korus2016evaluation}
Pawe{\l} Korus and Jiwu Huang,
\newblock ``Evaluation of random field models in multi-modal unsupervised tampering localization,''
\newblock in {\em IEEE WIFS}, 2016, pp. 1--6.

\bibitem{mayer2020exposing}
Owen Mayer and Matthew Stamm,
\newblock ``Exposing fake images with forensic similarity graphs,''
\newblock {\em IEEE JSTSP}, vol. 14, no. 5, pp. 1049--1064, 2020.

\bibitem{douglas2023data}
Benjamin Douglas, Patrick Ewell, and Markus Brauer,
\newblock ``Data quality in online human subjects research: Comparisons between mturk, prolific, cloudresearch, qualtrics, and sona,''
\newblock {\em Plos One}, vol. 18, 2023.

\bibitem{guan2019mfc}
Haiying Guan et~al.,
\newblock ``Mfc datasets: Large-scale benchmark datasets for media forensic challenge evaluation,''
\newblock in {\em IEEE/CVF WACV}, 2019, pp. 63--72.

\bibitem{novozamsky2020imd2020}
Adam Novozamsky, Babak Mahdian, and Stanislav Saic,
\newblock ``Imd2020: A large-scale annotated dataset tailored for detecting manipulated images,''
\newblock in {\em IEEE/CVF WACV}, 2020, pp. 71--80.

\bibitem{qin2020u2}
Xuebin Qin et~al.,
\newblock ``U2-net: Going deeper with nested u-structure for salient object detection,''
\newblock {\em Elsevier Pattern Recognition}, vol. 106, pp. 107404, 2020.

\bibitem{deng2018r3net}
Zijun Deng et~al.,
\newblock ``R3net: Recurrent residual refinement network for saliency detection,''
\newblock in {\em AAAI IJCAI}, 2018, pp. 684--690.

\bibitem{mejjati2020look}
Youssef Mejjati et~al.,
\newblock ``{Look here! A parametric learning based approach to redirect visual attention},''
\newblock in {\em Springer ECCV}, 2020, pp. 343--361.

\bibitem{chen2019guide}
Yen-Chung Chen et~al.,
\newblock ``Guide your eyes: Learning image manipulation under saliency guidance.,''
\newblock in {\em BMVC}, 2019, vol.~2, p.~3.

\bibitem{miangoleh2023realistic}
Mahdi Miangoleh et~al.,
\newblock ``Realistic saliency guided image enhancement,''
\newblock in {\em IEEE/CVF CVPR}, 2023, pp. 186--194.

\bibitem{agrawal2022sisl}
Susmit Agrawal et~al.,
\newblock ``Sisl: self-supervised image signature learning for splicing detection \& localization,''
\newblock in {\em IEEE/CVF CVPR}, 2022, pp. 22--32.

\bibitem{hessel2021clipscore}
Jack Hessel et~al.,
\newblock ``{CLIPS}core: A reference-free evaluation metric for image captioning,''
\newblock in {\em ACL EMNLP}, 2021, pp. 7514--7528.

\bibitem{takagi2022psm}
Yu~Takagi and Shinji Nishimoto,
\newblock ``High-resolution image reconstruction with latent diffusion models from human brain activity,''
\newblock in {\em IEEE/CVF CVPR}, 2022.

\end{thebibliography}

\end{document}